\documentclass{article} % For LaTeX2e
\usepackage{iclr2021_conference,times}

\usepackage{amsmath,amsfonts,bm}

\def\eqref#1{equation~\ref{#1}}
\def\1{\bm{1}}

\DeclareMathAlphabet{\mathsfit}{\encodingdefault}{\sfdefault}{m}{sl}
\SetMathAlphabet{\mathsfit}{bold}{\encodingdefault}{\sfdefault}{bx}{n}

\newcommand{\sigmoid}{\sigma}

\usepackage{enumitem}
\usepackage{hyperref}
\usepackage{url}
\usepackage{subcaption}
\usepackage{times}
\usepackage{epsfig}
\usepackage{graphicx}
\usepackage{amsmath}
\usepackage{amssymb}
\usepackage{xspace}
\usepackage{booktabs}
\usepackage[ruled]{algorithm2e}
\usepackage{algorithmic}
\SetKwComment{Comment}{$\triangleright$\ }{}
\usepackage{pifont}
\usepackage{xcolor}

\newcommand{\our}{AutoKD\xspace}

\title{AutoKD: Automatic Knowledge Distillation into a Student Architecture Family
}

\iclrfinalcopy

\author{Roy Henha Eyono$^1$ $^2$ \thanks{  Work done as an intern at Huawei Noah's Ark Lab (roy.eyono@mila.quebec) } , Fabio Maria Carlucci $^4$, Pedro M Esperança $^3$, Binxin Ru $^5$, Phillip Torr $^5$\\
$^1$McGill University, $^2$MILA, $^3$Huawei Noah’s Ark Lab, $^4$Facebook Research, $^5$University of Oxford \\
}

\begin{document}

\maketitle

\begin{abstract}
State-of-the-art results in deep learning have been improving steadily, in good part due to the use of larger models. However, widespread use is constrained by device hardware limitations, resulting in a substantial performance gap between state-of-the-art models and those that can be effectively deployed on small devices. While Knowledge Distillation (KD) theoretically enables small \textit{student} models to emulate larger \textit{teacher} models, in practice selecting a good student architecture requires considerable human expertise.
Neural Architecture Search (NAS) appears as a natural solution to this problem but most approaches can be inefficient, as most of the computation is spent comparing architectures \textit{sampled from the same distribution}, with negligible differences in performance. In this paper, we propose to instead search for a \textit{family} of student architectures sharing the property of being good at learning from a given teacher. Our approach \our, powered by Bayesian Optimization, explores a flexible graph-based search space, enabling us to automatically learn the optimal student architecture distribution and KD parameters, while being 20$\times$ more sample efficient compared to existing state-of-the-art. We evaluate our method on 3 datasets; on large images specifically, we reach the teacher performance while using 3$\times$ less memory and 10$\times$ less parameters. Finally, while \our uses the traditional KD loss, it outperforms more advanced KD variants using hand-designed students.
%Finally, we experimentally confirm that using KD can benefit NAS methods by significantly improving the correlation between different budgets in the context of multi-fidelity evaluations.
\end{abstract}

%%%%%%%%% BODY TEXT
% moved the overview at the end

\section{Introduction}

Recently-developed deep learning models have achieved remarkable performance in a variety of tasks.
However, breakthroughs leading to state-of-the-art (SOTA) results often rely on very large models: GPipe, Big Transfer and GPT-3 use $556$ million, $928$ million and $175$ billion parameters, respectively \citep{huang2019gpipe,Kolesnikov2020BiT,Brown2020GPT3}.

Deploying these models on user devices (e.g. smartphones) is currently impractical as they require large amounts of memory and computation; and even when large devices are an option (e.g. GPU clusters), the cost of large-scale deployment (e.g. continual inference) can be very high \citep{cheng2017survey}.
Additionally, target hardware does not always natively or efficiently support all operations used by SOTA architectures.
The applicability of these architectures is, therefore, severely limited, and workarounds using smaller or simplified models lead to a \textit{performance gap} between the technology available at the frontier of deep learning research and that usable in industry applications.

In order to bridge this gap, Knowledge Distillation (KD) emerges as a potential solution, allowing small \textit{student} models to learn from, and emulate the performance of, large \textit{teacher} models \citep{hinton2015KD}.
The student model can be constrained in its size and type of operations used, so that it will satisfy the requirements of the target computational environment.
Unfortunately, successfully achieving this 
%obtaining these improvements 
in practice is extremely challenging, requiring extensive human expertise.
For example, while we know that the architecture of the student is important for distillation \citep{liu2019search}, it remains unclear 
how to design the optimal network given some hardware constraints.
%how it should look, and also how to design the optimal network given some hardware constraints.

With Neural Architecture Search (NAS) it is possible to discover an optimal student architecture.
NAS automates the choice of neural network architecture for a specific task and dataset, given a \textit{search space} of architectures and a \textit{search strategy} to navigate that corresponding search space \citep{Pham2018_ENAS,Real2017_EvoNAS,Liu2019_DARTS,MANAS2019,Zela2018JAHS,ru2020NAGO}.
One important limitation of most NAS approaches is that the search space is very restricted, with a high proportion of resources spent on evaluating very similar architectures, thus rendering the approach limited in its effectiveness \citep{Yang2020NASEFH}. 
This is because traditional NAS approaches have no tools for distinguishing between architectures that are similar and architectures that are very different; as a consequence, computational resources are needed to compare even insignificant changes in the model (what we will call architectures within the same \textit{family} or sampled from the same \textit{distribution}).
Conversely, properly exploring a large space requires huge computational resources: for example, recent work by \cite{liu2019search} investigating how to find the optimal student requires evaluating $10,000$ models.

We propose an automated approach to knowledge distillation, in which we look for a \textit{family} of good students rather than a specific model. We find that even though our method, \our, does not output one specific architecture, all architectures sampled from the optimal \textit{family} of students perform well when trained with KD.
This reformulation of the NAS problem provides a more expressive search space containing very diverse architectures, thus increasing the effectiveness of the search procedure in finding good student networks. % without comprehensive human expertise.
In other words, by changing the focus from architecture search to architecture distribution search, we only evaluate meaningful architectural differences and, consequently, use computational resources more efficiently: evaluating 33$\times$ less architectures than comparable previous works \citep{liu2019search}, resulting in 20$\times$ more sample efficiency.

% Our contributions are as follows:
% \begin{itemize}[topsep=0pt,itemsep=2pt,partopsep=0pt, parsep=0pt]
%     \item A framework for combining KD with NAS and effectively emulate large models while using a fraction of the memory and of the parameters
%     \item By searching for the optimal student family, rather than for specific architectures, our algorithm is up to 20x more sample efficient than alternative NAS-based KD solutions.
%     \item We significantly outperform advanced KD methods on a benchmark of vision datasets, despite using the traditional KD loss, showcasing the efficacy of our found students.
% \end{itemize}

Our contributions are as follows: (\textbf{A}) a framework for combining KD with NAS and effectively emulate large models while using a fraction of the memory and of the parameters; (\textbf{B}) By searching for an optimal student family, rather than for specific architectures, our algorithm is up to 20x more sample efficient than alternative NAS-based KD solutions; (\textbf{C}) We significantly outperform advanced KD methods on a benchmark of vision datasets, despite using the traditional KD loss, showcasing the efficacy of our found students.

%\textcolor{red}{PEDRO SAYS: in the previous paragraphs we say <33X LESS EVALUATIONS> and <20X MORE SAMPLE EFFICIENT>. This will cause confusion.}

\section{Related Work}
Model compression has been studied since the beginning of the machine learning era, with multiple solutions being proposed \citep{choudhary2020comprehensive,cheng2017survey}.
Pruning based methods allow the removal of non-essential parameters from the model, with little-to-none drop in final performance. The primary motive of these approaches was to reduce the storage requirement, but they can also be used to speed up inference \citep{lecun1990optimal, han2015learning, li2016pruning}.
The idea behind quantization methods is to reduce the number of bits used to represent the weights and the activations in a model; depending on the specific implementation this can lead to reduced storage, reduced memory consumption and a general speed-up of the network \citep{fiesler1990weight,soudry2014expectation, rastegari2016xnor,zhu2016trained}.
In low rank factorization approaches, a given weight matrix is decomposed into the product of smaller ones, for example using singular value decomposition. When applied to fully connected layers this leads to reduced storage, while when applied to convolutional filters it leads to faster inference \citep{choudhary2020comprehensive}.

\begin{figure}[t!]
\centering
\includegraphics[width=0.82\textwidth]{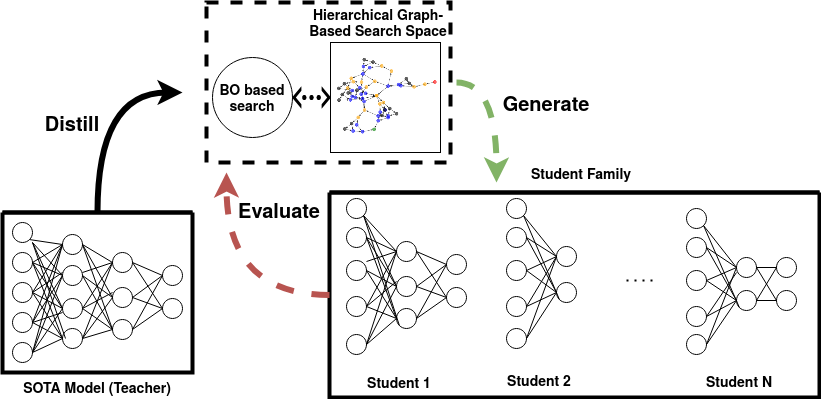} %,trim={0.1cm 19.1cm 2.1cm 3.2cm}, clip
\caption{
\our leverages multi-fidelity Bayesian Optimization, a hierarchical graph-based search space coupled with an architecture generator optimization pipeline, to find the optimal student for knowledge distillation. See section \ref{sec:methodology} for a detailed description.
% The \our framework combines Bayesian Optimization (BO), Neural Architecture Search (NAS) and Knowledge Distillation (KD). A hierarchical, graph-based search space defines a family of random network generators $G(\bm{\Theta})$ parameterized by a hyperparameter $\bm{\Theta}$, from where student networks are sampled.
% BO uses a surrogate model to propose generator hyperparameters, while students from these generators are trained with KD using a state-of-the-art teacher network. The student performances are evaluated and provided as feedback to update the BO surrogate model. To improve our BO surrogate model, the \textit{search} procedure is iterated, until the best family of student networks $G(\bm{\Theta}^*)$ is selected.
}
\label{fig:main_fig}
\end{figure}

All the above mentioned techniques can successfully reduce the complexity of a given model, but are not designed to substitute specific operations. For example, specialized hardware devices might only support a small subset of all the operations offered by modern deep learning frameworks. In Knowledge Distillation approaches, a large model (the teacher) distills its knowledge into a smaller student architecture \citep{hinton2015distilling}. This knowledge is assumed to be represented in the neural network's output distribution, hence in the standard KD framework, the output distribution of a student's network is optimized to match the teacher's output distribution for all the training data \citep{yun2020regularizing, ahn2019variational, yuan2020revisiting, tian2019contrastive, tung2019similarity}. 
%\RED{Talk a bit about some SOTA KD methods.}

The work of \citet{liu2019search} shows that the architecture of a student network is a contributing factor in its ability to learn from a given teacher. The authors propose combining KD with a traditional NAS pipeline, based on Reinforcement Learning, to find the optimal student. While this setup leads to good results, it does so at a huge computational cost, requiring over 5 days on 200 TPUs. 
Similarly, \cite{gu2020search} also look for the optimal student architecture, but do so by searching for a subgraph of the original teacher; therefore, it cannot be used to substitute unsupported operations.

Orthogonal approaches, looking at how KD can improve NAS, are explored by
\cite{trofimov2020multi} and \cite{li2020block}. The first establishes that KD improves the correlation between different budgets in multi-fidelity methods, while the second uses the teacher supervision to search the architecture in a blockwise fashion.

\citet{stanton2021does} argue that SOTA KD methods successfully improve student \textit{generalization}, the performance of a student in predicting unseen data, but struggles to obtain good student \textit{fidelity}, the ability of a student to match teacher predictions. With \our, we successfully produce high fidelity student families using a fraction of the parameters.

% Current NAS applications for KD adopt rigid search space design practices. \citet{liu2019search} do not optimize KD temperature and weight, why?
% Their search space is quite rigid: it consists of seven predefined blocks, with each block contains a list of identical layers. Our search space allows us to search for the number of layers, the convolution and skip operation type, conv kernel size, squeeze-and-excite ratio, and input/output filter size, for each block independently.
% They also use a weighted product of accuracy and latency (fixed pareto front, not true multi-objective). They are very sample inefficient, they sample $10k$ models and train all of them for 5 epochs on ImageNet. Top models (how many?) are then trained for another 400 epochs.

% \textbf{Ours:} We sample end-to-end architectures from our generator with minimal expert insight for neural architecture optimization.

% ...

% Things to mention:
% - why NAS is inefficient
% - why GAS is better
% - why we ignore one-shot methods (maybe we could add this in related work)
% - example of how memory constrained phones need small networks

% ...

% ``A Survey of Model Compression and Acceleration for Deep Neural Networks'', https://arxiv.org/pdf/1710.09282.pdf

\section{Searching for the optimal student network generator}
\label{sec:methodology}

% \textbf{Teacher assistant.} It has been shown \citep{mirzadeh2019improved} that KD can fail when the teacher model has much more capacity than the student; to solve this an intermediate model, the teacher assistant, was introduced \citep{mirzadeh2019improved}.
% We are the first to search for the student and the teacher assistant at the same time \textbf{TODO}.

% \textbf{Feature representation similarity}. Similarly to \cite{teacherGuidedNAS} we compute a similarity of feature similarity between the teacher and the student and use this score to select the best students. \textbf{TODO}

The \our framework (Fig.~\ref{fig:main_fig}) combines Bayesian Optimization (BO), Neural Architecture Search (NAS) and Knowledge Distillation (KD). \our defines a family of random network generators $G(\bm{\theta})$ parameterized by a hyperparameter $\bm{\theta}$, from where student networks are sampled.
BO uses a surrogate model to propose generator hyperparameters, while students from these generators are trained with KD using a state-of-the-art teacher network. The student performances are evaluated and provided as feedback to update the BO surrogate model. To improve our BO surrogate model, the \textit{search} procedure is iterated, until the best family of student networks $G(\bm{\theta}^*)$ is selected.
% \our leverages multi-fidelity Bayesian Optimization (BOHB) \citep{falkner2018bohb}; a hierarchical, graph-based search space and the architecture generator optimization of NAGO \citep{ru2020NAGO}; to deal with Knowledge Distillation \citep{hinton2015distilling}.
% In this section we specify all components of \our.
In this section we specify all components of \our. See also Fig.~\ref{fig:main_fig} for an overview.

\subsection{Knowledge Distillation}

%\begin{itemize}
   % \item Define Knowledge Distillation
    %\item Show KD loss equation here...
    %\item Cite relevant literature
    %\item Note: Discuss mainly traditional Knowledge Distillation method, give variants an honourable mention.
%\end{itemize}

Knowledge Distillation (KD; \citealp{hinton2015distilling}) is a method to transfer, or distill, knowledge from one model to another---usually from a large model to small one---such that the small \textit{student} model learns to emulate the performance of the large \textit{teacher} model.
KD can be formalized as minimizing the objective function:
\begin{equation}
    \mathcal{L}_{\text{KD}} = \sum_{x_{i} \in X} l(f_{T}(x_i), f_{S}(x_i))
    \label{eqn:KDLoss}
\end{equation}
where $l$ is the loss function that measures the difference in performance between the teacher $f_{T}$ and the student $f_{S}$, and $x_i$ is the $i$th input. 
%$y_i$ is the $i$th target. 
The conventional loss function $l$ used in practice is a linear combination of the traditional cross entropy loss $L_{\text{CE}}$ and the Kullback--Leibler divergence $L_{\text{KL}}$ of the pre-softmax outputs for $f_{T}$ and $f_{S}$:
\begin{equation}
    l = (1-\alpha)L_{\text{CE}} +  \alpha L_{\text{KL}}\left( \sigmoid \left( {f_{T}(x_i)}/{\tau} \right), \sigmoid \left( {f_{S}(x_i)}/{\tau} \right) \right)
    \label{eqn:HintonLoss}
\end{equation}
where $\sigmoid$ is the softmax function $\sigmoid(x) = 1/(1 + \exp(-x))$, and $\tau$ is the softmax temperature. \citet{hinton2015distilling} propose ``softening" the probabilities using temperature scaling with $\tau \ge 1$. The parameter $\alpha$ represents the weight trade-off between the KL loss and the cross entropy loss $L_{\text{CE}}$. The $\mathcal{L}_{\text{KD}}$ loss is characterized by the hyper-parameters: $\alpha$ and $\tau$; popular choices are $\tau \in \{3,4,5\}$ and $\alpha = 0.9$ \citep{huang2017like,zagoruyko2016paying, zhu2018knowledge}. Numerous other methods \citep{polino2018model, huang2017like, tung2019similarity} can be formulated as a form of Equation (\ref{eqn:HintonLoss}), but in this paper we use the conventional loss function $l$.

Traditionally in KD, both the teacher and the student network have predefined architectures.
In contrast, \our defines a search space of student network architectures and finds the optimal student by leveraging neural architecture search, as detailed below.

\begin{algorithm} %TODO multiobjective or BOHB?!
[tb]
   \caption{AutoKD}
   \label{alg:autokd}
\begin{algorithmic}[1]
   \STATE {\bfseries Input:} Network generator $G$, BOHB hyperparameters($\eta$, training budget $b_{min}$ and $b_{max}$), Evaluation function $f_{KD}(\boldsymbol{\theta}, b)$ which assesses the validation performance of a generator hyperparameter $\boldsymbol{\theta}$ by sampling an architecture from $G(\boldsymbol{\theta})$ and training it with the KD loss $\mathcal{L}_{\text{KD}}$ (equations \ref{eqn:KDLoss} and \ref{eqn:HintonLoss}) for $b$ epochs.
   \STATE $s_{max} = \lfloor \log_{\eta} \frac{b_{max}}{b_{min}} \rfloor$;
%   \REPEAT
   \FOR{ $s \in \{s_{max}, s_{max}-1, \dots, 0 \}$}
        \STATE Sample $M = \lceil \frac{s_{max}+1}{s+1}\cdot \eta^s \rceil$ generator hyperparameters $\boldsymbol{\Theta} = \{\boldsymbol{\theta}^j\}_{j=1}^M$ which maximises the raito of kernel density estimators \Comment*[r]{\cite[Algorithm 2]{falkner2018bohb}} 
        \STATE Initialise $b = \eta^s \cdot b_{max}$ \Comment*[r]{Run Successive Halving \citep{li2016hyperband}}
        \WHILE{$b \leq b_{max}$}
        \STATE $\mathbf{L} = \{f_{KD}(\boldsymbol{\theta}, b): \boldsymbol{\theta} \in \boldsymbol{\Theta} \}$;
        \STATE $\boldsymbol{\Theta} = top\_k (\boldsymbol{\Theta}, \mathbf{L}, \lfloor |\boldsymbol{\Theta}|/\eta \rfloor)$;
        \STATE $b = \eta \cdot b$;
        \ENDWHILE
   \ENDFOR
   \STATE Obtain the best performing configuration $\boldsymbol{\theta}^*$ for the student network generator.
   %or the Pareto set $\boldsymbol{\Theta}^*$
   \STATE Sample $k$ architectures from $G(\boldsymbol{\theta}^*)$, train them to completion, and obtain test performance.
\end{algorithmic}
\end{algorithm}

\subsection{Student search via Generator Optimization}\label{sec:nago}

Most NAS method for vision tasks employ a cell-based search space, where networks are built by stacking building blocks (\textit{cells}) and the operations inside the cell are searched \citep{Pham2018_ENAS,Real2017_EvoNAS,Liu2019_DARTS}.
This results in a single architecture being output by the NAS procedure.
In contrast, more flexible search spaces have recently been proposed that are based on neural network generators \citep{xie2019exploring, ru2020NAGO}.
The generator hyperparameters define the characteristics of the family of networks being generated.
%, and optimizing these hyperparameters gives an optimal \textit{family} of architectures that perform well on the given task.

NAGO \citep{ru2020NAGO} optimizes an architecture generator instead of a single architecture and proposes a hierarchical graph-based space which is highly expressive yet low-dimensional. Specifically, the search space of NAGO comprises three levels of graphs (where the node in the higher level is a lower-level graph). The top level is a graph of cells ($G_{top}$) and each cell is itself a graph of middle-level modules ($G_{mid}$). Each module further corresponds to a graph of bottom-level operation units ($G_{bottom}$) such as a relu-conv3$\times$3-bn triplet. NAGO adopts three random graph generators to define the connectivity/topology of $G_{top}$, $G_{mid}$ and $G_{bottom}$ respectively, and thus is able to produce a wide variety of architectures with only a few generator hyperparameters.
% The \our framework follows the same approach as NAGO.
% \our extends NAGO to the KD setting.
% \our builds on NAGO's three layer graph-based search space for knowledge distillation.
\our employs NAGO as the NAS backbone for finding the optimal student family.

Our pipeline consists of two phases. In the first phase (\textit{search}), a multi-fidelity Bayesian optimisation technique, BOHB \citep{falkner2018bohb}, is employed to optimise the low-dimensional search space.
BOHB uses partial evaluations with smaller-than-full budget to exclude bad configurations early in the search process, thus saving resources to evaluate more promising configurations. Given the same
time constraint, BOHB evaluates many more configurations than conventional BO which evaluates all configurations with full budget. 
As \citet{ru2020NAGO} empirically observe that good generator hyperparameters lead to a tight distribution of well-performing architectures (small performance standard deviation), we similarly assess the performance of a particular generator hyperparameter value with only one architecture sample.
In the second phase (\textit{retrain}), \our uniformly samples multiple architectures from the optimal generator found during the search phase and evaluates them with longer training budgets to obtain the best architecture performance. 

Instead of the traditionally used cross-entropy loss, \our uses the KD loss in \eqref{eqn:HintonLoss} to allow the sampled architecture to distill knowledge from its teacher.
The KD hyperparameters temperature $\tau$ and loss weight $\alpha$ are included in the search space and optimized simultaneously with the architecture to ensure that the student architectures can efficiently distill knowledge both from the designated teacher and the data distribution. A full overview of the framework is shown in Fig. \ref{fig:main_fig} and Algorithm \ref{alg:autokd}.
%We show that we can improve NAGO with KD in both the search and the retraining phase.

\section{Experiments}
\begin{figure}[h!]
    \centering
    \includegraphics[width=1\textwidth]{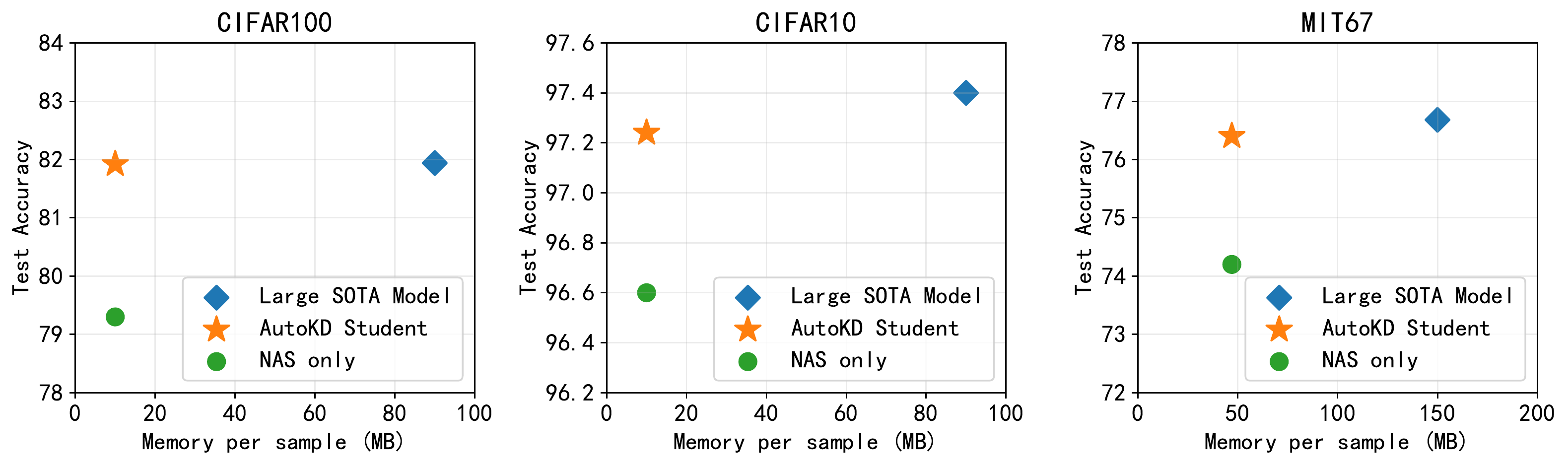}
    \vspace{-0.7cm}
    \caption{Accuracy vs memory per sample for the SOTA Model (the teacher), the \our student and best architecture found by vanilla NAS. Note that the MIT67 teacher has almost 10$\times$ the number of parameters of the student (54M vs 6M).
    %\RED{REDO PLOTS: x-axis says params but should be ``memory (megabytes per sample)'', y-axis should say ``validation/test accuracy (\%)'', consistent dataset name spelling in the title (we're using uppercase in the text)}
    }
    \label{fig:acc_vs_mem}
\end{figure}

The first part of this section studies how KD can improve the performance of our chosen NAS backbone (NAGO). In the second part, we show how a family of students, when trained with KD (\our), can emulate much larger teachers, significantly outperforming current hand-crafted architectures.

\textbf{Experimental setup.} All of our experiments were run on the two, small-image, standard object recognition datasets \textit{CIFAR10} and \textit{CIFAR100} \citep{Krizhevsky2009_cifar}, as well as \textit{MIT67} for large-image scene recognition \citep{quattoni2009_mit67}.
We limit the number of student network parameters to $4.0$M for small-image tasks and $6.0$M for large-image tasks. Following \cite{liu2019search}, we picked Inception-Resnet-V2 \citep{szegedy2016inception} as a teacher for the large image dataset. As that model could not be directly applied to small images, and to explore the use of a machine-designed network as a teacher, we decided to use the best DARTS \citep{Liu2019_DARTS} architecture to guide the search on the CIFAR datasets. 
For \textit{ImageNet} \citep{Deng2009_imagenet}, we use a Inception-Resnet-V2 teacher.
All experiments are run on NVIDIA Tesla V100 GPUs.

% \textbf{Datasets} In KD-NAGO, we distill knowledge from an expert teacher model to a family of lightweight student models to address limitations with low-resource device. To demonstrate our method, we perform experiments across different image recognition datasets, namely, CIFAR10, CIFAR100, SPORT8 and MIT67. In our experiments, for CIFAR10 and CIFAR100, we adopt an image size of 32$\times$32 and for SPORT8 and MIT67, an image size of ?$\times$?. 

% \textbf{Teachers} For each dataset, we use specialized teacher models that are optimized for each respective dataset. We leverage the Differential Architecture Search (DARTS) method to search for optimized architectures for CIFAR10 and CIFAR100, and for MIT67 and Sport8, we use the InceptionResNetV2 model as a teacher. Our teachers report SOTA results across our image recognition datasets:

% \begin{itemize}
%     \item \textbf{CIFAR10} - $97.34\%$ Top-1 accuracy; Teacher: DARTS: approx. 6M parameters
%     \item \textbf{CIFAR100} - Teacher: DARTS: approx. 6M parameters
%     \item \textbf{Sport8} - Teacher: InceptionResNetv2: 56M parameters
%     \item \textbf{MIT67} - Teacher: InceptionResNetv2: 56M parameters
% \end{itemize}

% Our choice of teacher models was motivated by their empirical performances, specifically each teacher outperforms the NAGO benchmark across our pre-defined datasets which is a necessary requirement for us to compare the empirical gains of our NAGO students. 

\textbf{NAS implementation. }
Our approach follows the search space and BO-based search protocol proposed by NAGO \citep{ru2020NAGO}, as such our student architectures are based on hierarchical random graphs. Likewise, we employ a multi-fidelity evaluation scheme based on BOHB \citep{falkner2018bohb} where candidates are trained for different epochs ($30$, $60$ and $120$) and then evaluated on the validation set. In total, only $\sim$300 models are trained during the search procedure: using 8 GPUs, this amounts to $\sim$2.5 days of compute on the considered datasets.
At the end of the search, we sample $8$ architectures from the best found generator, train them for 600 epochs (with KD, using the optimal temperature and loss weight found during the search), and report the average performance (top-1 test accuracy). All remaining training parameters were set following \cite{ru2020NAGO}.

In \our, we include the knowledge distillation hyperparameters, \textit{temperature} and \textit{weight}, in the search space, so that they are optimized alongside the architecture. 
The {temperature} ranges from $1$ to $10$, while the {weight} ranges from $0$ to $1$.

% In the endFollowing \cite{ru2020NAGO}, we sample a collection of architecture generators and initially evaluate them for $30$ epochs. The top $20$--$30\%$ best performing generators progress to the subsequent budget of $60$ epochs, where the generator is re-evaluated amongst additional generator random samples. This procedure is repeated at the final budget of $120$ epochs.

\subsection{Impact of Knowledge Distillation on NAS}
\begin{figure}[!ht]
\centering
\begin{tabular}{ccc}
\fbox{\includegraphics[width=4cm]{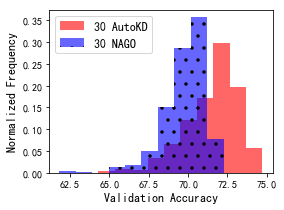}}&
\fbox{\includegraphics[width=4cm]{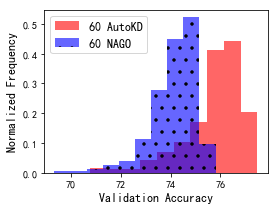}}&
\fbox{\includegraphics[width=4cm]{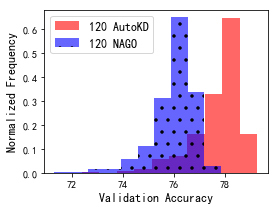}}\\
(a) 30 epochs&(b) 60 epochs&(c) 120 epochs
\end{tabular}
\vspace{0.2cm}
\caption{Top-1 accuracy distribution for \our and standard NAGO at different budgets on CIFAR100. The histograms are tallied across $5$ runs. Across all budgets, \our samples architectures with improved performances in top-1 accuracy compared to NAGO.}
\label{fig:acc_dist}
\end{figure}

\begin{figure}[!ht]
\centering
\begin{tabular}{ccc}
\fbox{\includegraphics[width=4cm]{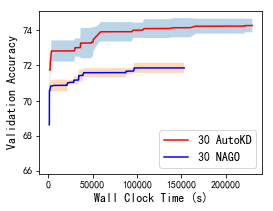}}&
\fbox{\includegraphics[width=4cm]{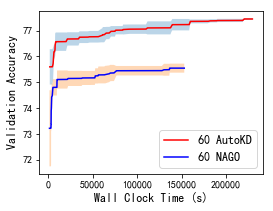}}&
\fbox{\includegraphics[width=4cm]{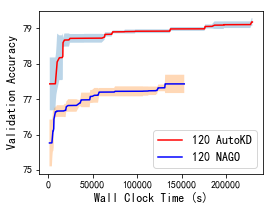}}\\
(a) 30 epochs&(b) 60 epochs&(c) 120 epochs
\end{tabular}
\vspace{0.25cm}
\caption{Top-1 accuracy of the best model found during search at a given computation time on CIFAR100 for \our (red) and NAGO (blue) across different budgets. Each method was run 8 times with the bold curve showing the average performance and the shaded region the stdev. }
\label{fig:acc_plot}
\end{figure}

To understand the contribution from KD, we first compare vanilla NAGO with \our on CIFAR100. Fig.~\ref{fig:acc_dist} shows  the validation accuracy distribution at different epochs: clearly, using KD leads to better performing models. 
Indeed this can be seen in more detail in Fig.~\ref{fig:acc_plot}, where we show the performance of the best found model vs the wall clock time for each budget. It is worth mentioning that while the KD version takes longer (as it needs to compute the lessons on the fly), it consistently outperforms vanilla NAGO by a significant margin on all three datasets.

Note that accuracies in Fig.~\ref{fig:acc_plot} refer to the best models found during the search process, while Fig.~\ref{fig:acc_dist} shows the histograms of all models evaluated during search, which are by definition lower in accuracy, on average. At the end of search, the model is retrained for longer (as commonly done in NAS methods), thus leading to the higher accuracies also shown in Fig. \ref{fig:training_curves}.

\begin{figure}[!ht]
\centering
\begin{tabular}{cc}
\fbox{\includegraphics[trim={0cm 0cm 0cm 0.2cm},clip,width=4cm]{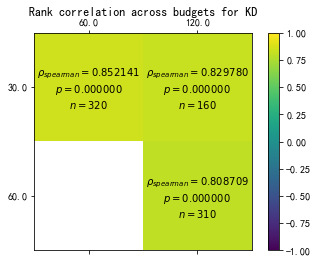}}&
\fbox{\includegraphics[trim={0cm 0cm 0cm 0.2cm},clip,width=4cm]{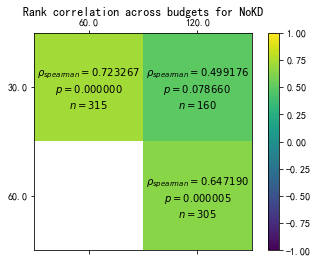}}
\end{tabular}
\vspace{0.2cm}
\caption{Rank Correlations between different epoch budgets for \our (KD; left) and Standard NAGO (right) computed for 5 runs of NAGO and \our respectively. NAGO reports a rank correlation coefficient of $5 \cdot 10^{-1}$ for epoch pair 30-120, which is $3.3 \cdot 10^{-1}$ less than that of the KD rank correlation. These results show that the rank correlation across all budget pairs vastly improves when knowledge distillation is applied.}
\label{fig:rank_correlation}
\end{figure}

Not only does \our offer better absolute performance, but it also enables better multi-fidelity correlation, as can be seen in Fig.~\ref{fig:rank_correlation}. 
For example, the correlation between $30$ and $120$ epochs improves from $0.49$ to $0.82$ by using KD, a result that is consistent with the findings in \cite{trofimov2020multi}. 
Note that multi-fidelity methods work under the assumption that the rankings at different budgets remains consistent to guarantee that the best models progress to the next stage. A high correlation between the rankings is, as such, crucial.

% sample several architectures from a fixed CIFAR10 NAGO generator  and compare across various weight and temperature parameters and report our findings. 

% In Fig. \ref{fig:ablation_study}, we train each sampled architecture for 600 epochs with an expert DARTS teacher trained on the CIFAR10 dataset. In our study, we observe a strong positive correlation between the weight parameter in KD and the empirical performance of the sampled architecture. It appears that sampled architectures that perform well on CIFAR-10 are those that maximize on knowledge from the DARTS teacher. This experiment serves as an illustration of the importance of choosing the right \textit{temperature} and \textit{weight} hyperparameters.

\begin{figure}[!ht]
\centering
\begin{tabular}{cc}
\fbox{\includegraphics[trim=2.9cm 1.8cm 2.2cm 2cm, clip,width=4cm]{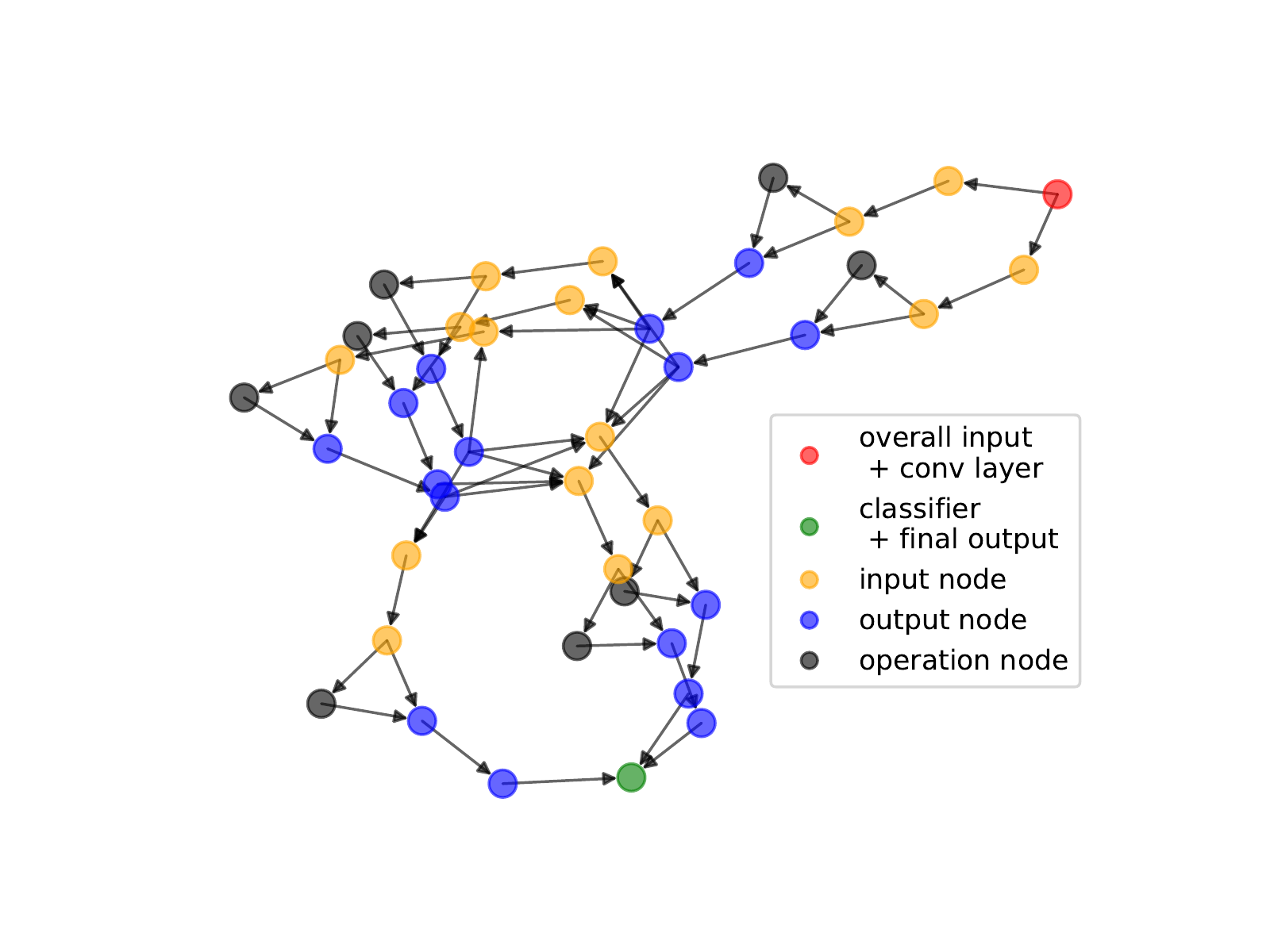}}&
\fbox{\includegraphics[trim=2.9cm 1.8cm 2.2cm 2cm, clip, width=4cm]{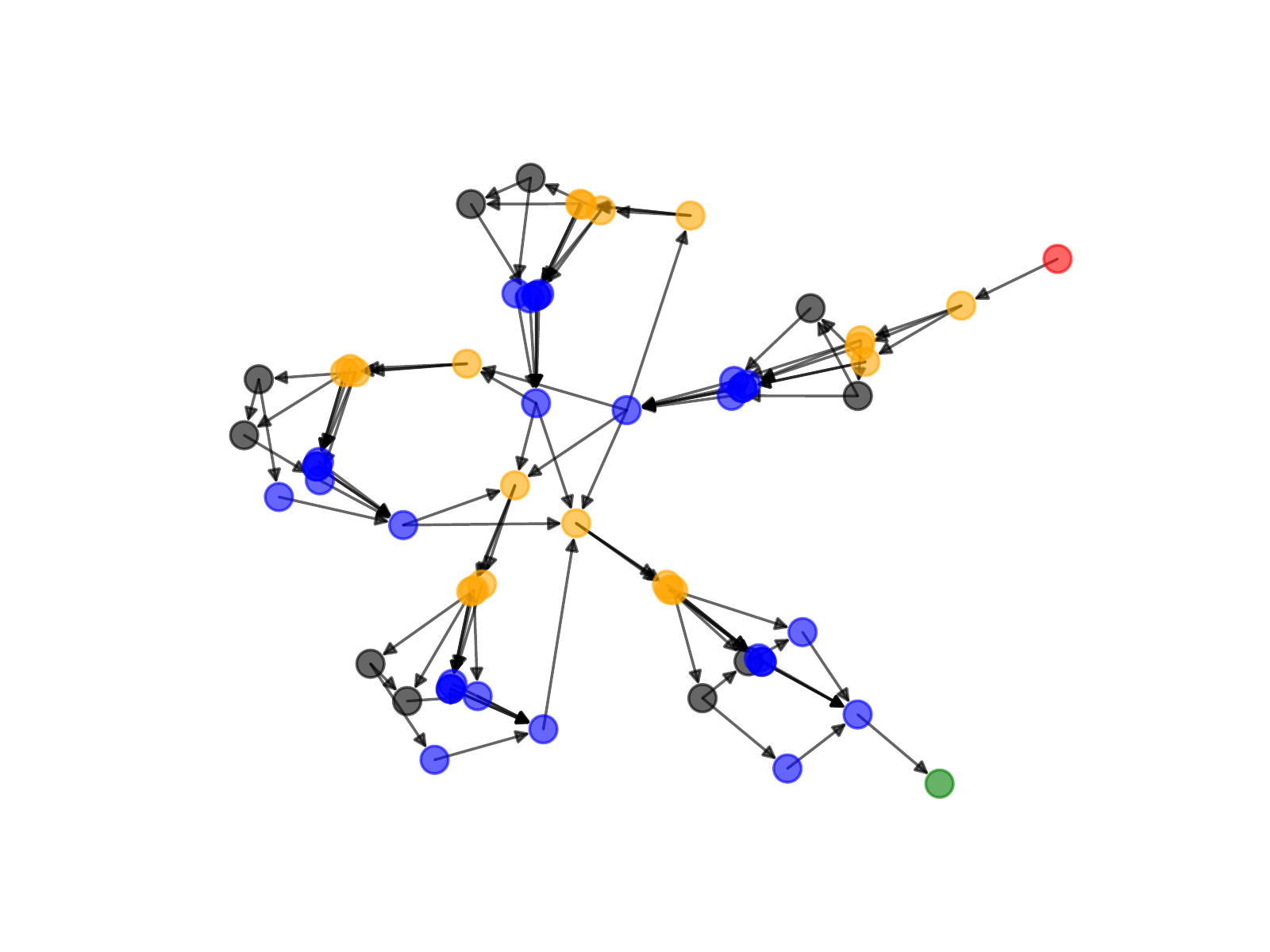}}
\end{tabular}
\vspace{0.2cm}
\caption{Networks sampled from the best generator parameters found by \our (left) and NAGO (right) on CIFAR10. The former is organized with 8 clusters of 3 nodes, while the latter has 5 clusters of 10 nodes, showcasing how the optimal configuration depends on teacher supervision.}
\label{fig:cifar10_graphs}
\end{figure}

\subsection{Large model emulation}
\begin{figure}[!ht]
\centering
\begin{tabular}{ccc}
\fbox{\includegraphics[width=4cm]{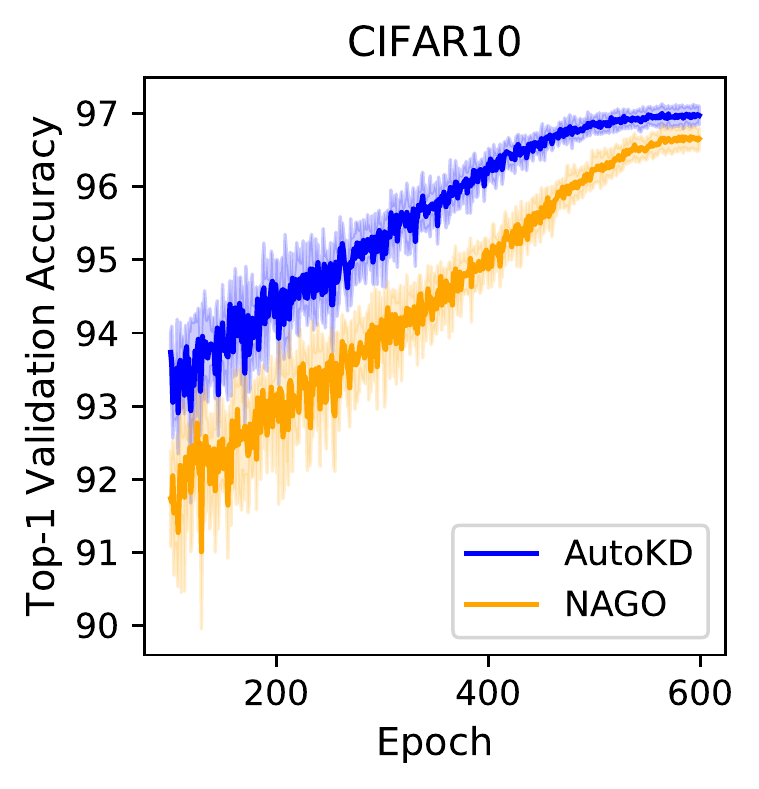}}&
\fbox{\includegraphics[width=4cm]{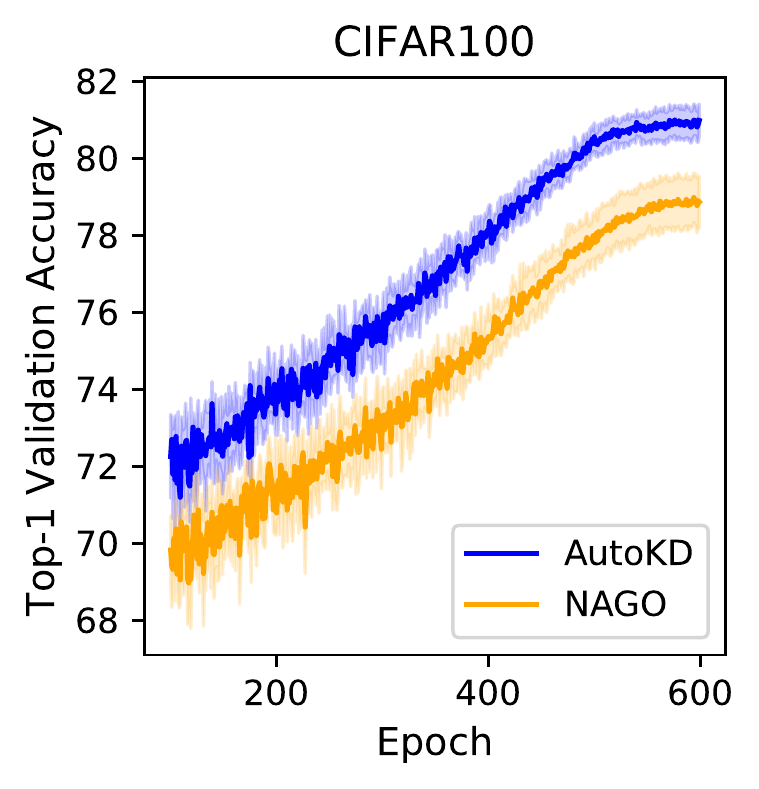}}&
\fbox{\includegraphics[width=4cm]{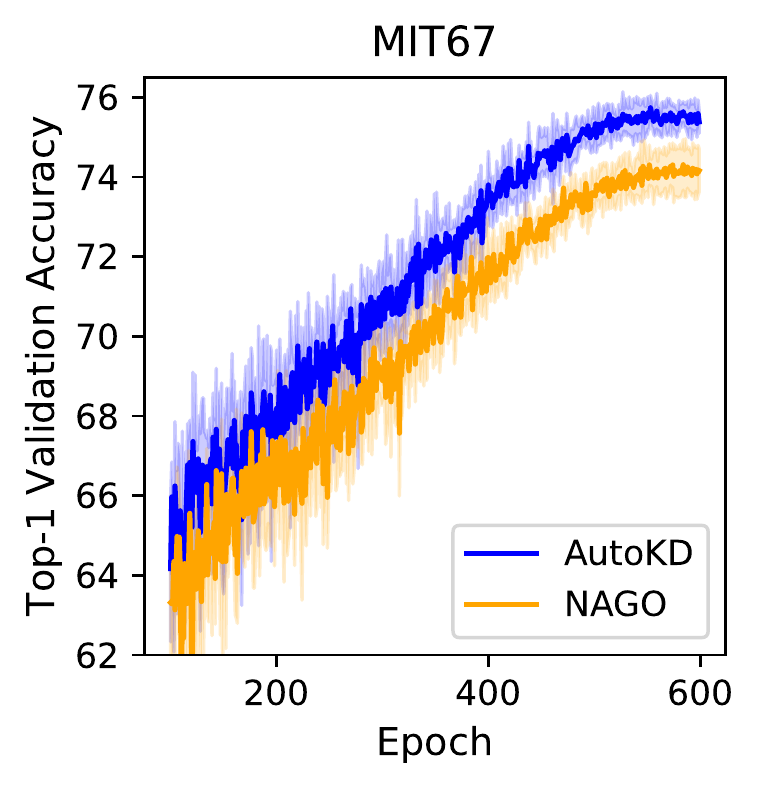}}\\
\end{tabular}
\vspace{0.01cm}
\caption{Final training curves for the top generator found by NAGO and \our, for CIFAR10, CIFAR100 and MIT67. 
Each generator was sampled 8 times and the 8 corresponding architectures trained for 600 epochs. Bold line represents the average; shaded region represents std deviation.}
\label{fig:training_curves}
\end{figure}

At its core, \our's goal is to emulate the performance of large SOTA models with smaller students. Fig.~\ref{fig:acc_vs_mem} shows how the proposed method manages to reach the teacher's performance while using only 1/9th of the memory on small image datasets. On MIT67, the found architecture is not only using 1/3rd of the memory, but also 1/10th of parameters. Finally, it is worth noting how \our increases student performance, as such the high final accuracy cannot only be explained by the NAS procedure.  
Indeed, looking at Fig.~\ref{fig:training_curves} it is clear how KD improves both the speed of convergence and the final accuracy. Furthermore, as shown in Fig. \ref{fig:cifar10_graphs}, the optimal family of architectures is actually different when searched with KD

\newcommand{\up}[1]{\textcolor{red}{$\bm{\uparrow}$}$\times #1$}
\newcommand{\down}[1]{\textcolor{green}{$\bm{\downarrow}$}$\times #1$}
\begin{table}[ht!]
\caption{Comparison with KD state-of-the-art. \our uses the standard KD loss \citep{hinton2015distilling}, while competing methods are using modern variants. 
In column ``S acc'', the improvement of the student accuracy with KD versus the same student without KD is given in parenthesis (+v).
``\up{f}'' and ``\down{f}'' denotes the increase/decrease in parameter count by the factor $f$ relative to \our for the same dataset. 
The top performing student accuracy (\emph{S acc}) for each dataset is specified in bold. 
For each dataset, we sampled 8 architectures and averaged them over 5 runs. For MIT67, the VID method is a transfer learning task from ImageNet to MIT67, hence the absence of teacher accuracy (\emph{T acc}) statistics.}
\label{tab:kd}
\begin{center}
\begin{small}
\begin{tabular*}{\textwidth}{@{}llllll@{}}
\toprule
Method$\dagger$ & Teacher (T) & Student (S)  & S params  & T acc & S acc \\ \midrule
\multicolumn{6}{c}{\textbf{ImageNet}}\\
\midrule
     \textbf{AKD}  &   InceptionResNetV2   &  AKDNet &   ---  & 75.5 & 75.5 (+2.5)   \\
     \textbf{CRD}  &   ResNet-34   &  ResNet-18 &   11.5M \up{1.9}  & 73.3 & 71.4 (+1.6)   \\
     \textbf{KD-LSR}  &   ResNet-50   &  ResNet-50 &   23M \up{3.8}  & 75.8 & 76.4 (+0.6)   \\
     \midrule
     \midrule
     \textbf{\our} (ours)  &  InceptionResNetV2 & NAGO &   6.0M  & 75.5  & \textbf{78.0} (+1.2)   \\
     \midrule
     \midrule
\multicolumn{6}{c}{\textbf{MIT67}}\\
\midrule
     \textbf{SKD}  &  ResNet-18   & ResNet-18 &   11.5M \up{1.9}  & 55.3 & 60.4 (+5.1)     \\
   \textbf{VID}  &  ResNet-34   & ResNet-18      &     11.5M \up{1.9}  & --- & 71.9 (+0.9) \\
     \textbf{VID}  &  ResNet-34   & VGG-9        & 10.9M \up{1.8}  & --- &   72.0 (+6.0)   \\
     \midrule
     \midrule
     \textbf{\our} (ours)  &  InceptionResNetV2 & NAGO &   6.0M  & 76.6  & \textbf{76.0} (+1.8)   \\
     \midrule
     \midrule
\multicolumn{6}{c}{\textbf{CIFAR100}}\\
\midrule
   
     \textbf{CRD}  &  ResNet-32 $\times$ 4   & ShuffleNetV2  &   7.4M \up{1.9}  & 79.4 & 75.7 (+3.8)     \\
     \textbf{CRD}  &  WRN 40-2  & WRN-16-2  &   0.7M \down{5.7}  & 75.6  &  75.6 (+2.4)    \\
     \textbf{VID}  &  WRN 40-2   & WRN 40-2      &     2.2M \down{1.8}  & 74.2 & 76.1 (+1.8)  \\
     \textbf{KD-LSR}  &  ResNet-18   & ResNet-18 &   11.5M \up{2.9}   & 75.9 & 77.4 (+1.5)     \\
     \textbf{SKD}  &  ResNet-18   & ResNet-18 &   11.5M \up{2.9} &    75.3 & 79.6 (+4.3)      \\
     \textbf{KD-LSR}  &  DenseNet-121   & DenseNet-121 &   7.0M \up{1.8}  & 79.0 & 80.3 (+1.3)      \\
     \midrule
     \midrule
     \textbf{\our} (ours)  &  DARTS & NAGO &   4.0M &    81.9 & \textbf{81.2} (+2.6)     \\
     \midrule
     \midrule
%\multicolumn{6}{c}{\textbf{CIFAR10}}\\
%\midrule
%   \textbf{VID}  &  WRN 40-2   & WRN 16-1      &     0.7M \down{5.7}  & 94.3 & 91.9 (+1.2) \\
%   \textbf{SPKD}  &  WRN 16-8   & WRN 40-2      &     2.2M \down{1.8}  & 95.8 & 95.5 (+0.6) \\
%   \textbf{AT}  &  WRN 16-8   & WRN 40-2      &     2.2M \down{1.8}  & 95.8 & 95.5 (+0.6) \\
%     \midrule
%     \midrule
%     \textbf{\our} (ours)  &  DARTS & NAGO &   4.0M  & 97.4 &   \textbf{97.1} (+0.8)     \\
%     \midrule
%     \midrule
\end{tabular*}
\end{small}
\end{center}
\small{$\dagger$ AKD \citep{liu2019search}, SKD \citep{yun2020regularizing}, VID \citep{ahn2019variational}, KD-LSR \citep{yuan2020revisiting}, CRD \citep{tian2019contrastive}}%, SPKD \citep{tung2019similarity}, AT \citep{zagoruyko2016paying}.}
\end{table}

\textit{ImageNet, MIT67, CIFAR100.} 
Table~\ref{tab:kd} shows the comparison of \our with other KD methods.
Notice how learning the student architecture allows \our to outperform a variety of more advanced KD approaches while emplying a smaller parameter count in the student. 
Finally, \our is orthogonal to advanced KD approaches and could be combined with any of them for even further increases in performance.

The improved results on smaller datasets extend to large datasets as well. On ImageNet, AutoKD reaches $78.0$\% top-1 accuracy, outperforming both \cite{liu2019search} using the same teacher ($75.5$\%) and vanilla NAGO ($76.8$\%).

\section{Discussion and Conclusion}
Improving Knowledge Distillation by searching for the optimal student architecture is a promising idea that has recently started to gain attention in the community \citep{liu2019search,trofimov2020multi,gu2020search}.
In contrast with earlier KD-NAS approaches, which search for specific architectures, our method searches for a \textit{family} of networks sharing the same characteristics.
The main benefit of this approach is sample efficiency: while traditional methods spend many computational resources evaluating similar architectures \citep{Yang2020NASEFH}, \our is able to avoid this pitfall:
for instance, the method of \cite{liu2019search} requires $\sim \! 10,000$ architecture samples, while \our can effectively search for the optimal student family with only $300$ samples.
Compared to traditional KD methods, \our is capable of achieving better performance with student architectures that have less parameters (and/or use less memory) than hand-defined ones.

AutoKD demonstrates the fact that the macro-structure (connectivity and capacity) of a network is more important than its micro-structure (the specific operations). This has been shown to be true for non-KD NAS \citep{xie2019exploring,ru2020NAGO} and is here experimentally confirmed for KD-NAS as well.
Changing the focus of optimization in this way releases computational resources that can be used to effectively optimize the global properties of the network. %, such as the proportion of different types of convolutions and the degree of connectivity.
Additionally, the fact that a family of architectures can characterized by a small number of hyperparameters makes the comparison of architectures more meaningful and interpretable. In the current implementation, \our finds the optimal student family, in which all sampled architectures perform well: future work should explore how to fully exploit this distribution, possibly finetuning the network distribution to obtain an ever better performing model.

To summarize, \our offers a strategy to efficiently emulate large, state-of-the-art models with a fraction of the model size. Indeed, our family of searched students consistently outperforms the best hand-crafted students on CIFAR10, CIFAR100 and MIT67.

% \section{Overview - this section is temporary}
% \input{overview}

\bibliography{refs}
\bibliographystyle{iclr2021_conference}

\appendix
\newpage
\section{Appendix}
\begin{figure}[!h]
\centering
\includegraphics[width=0.5\textwidth]{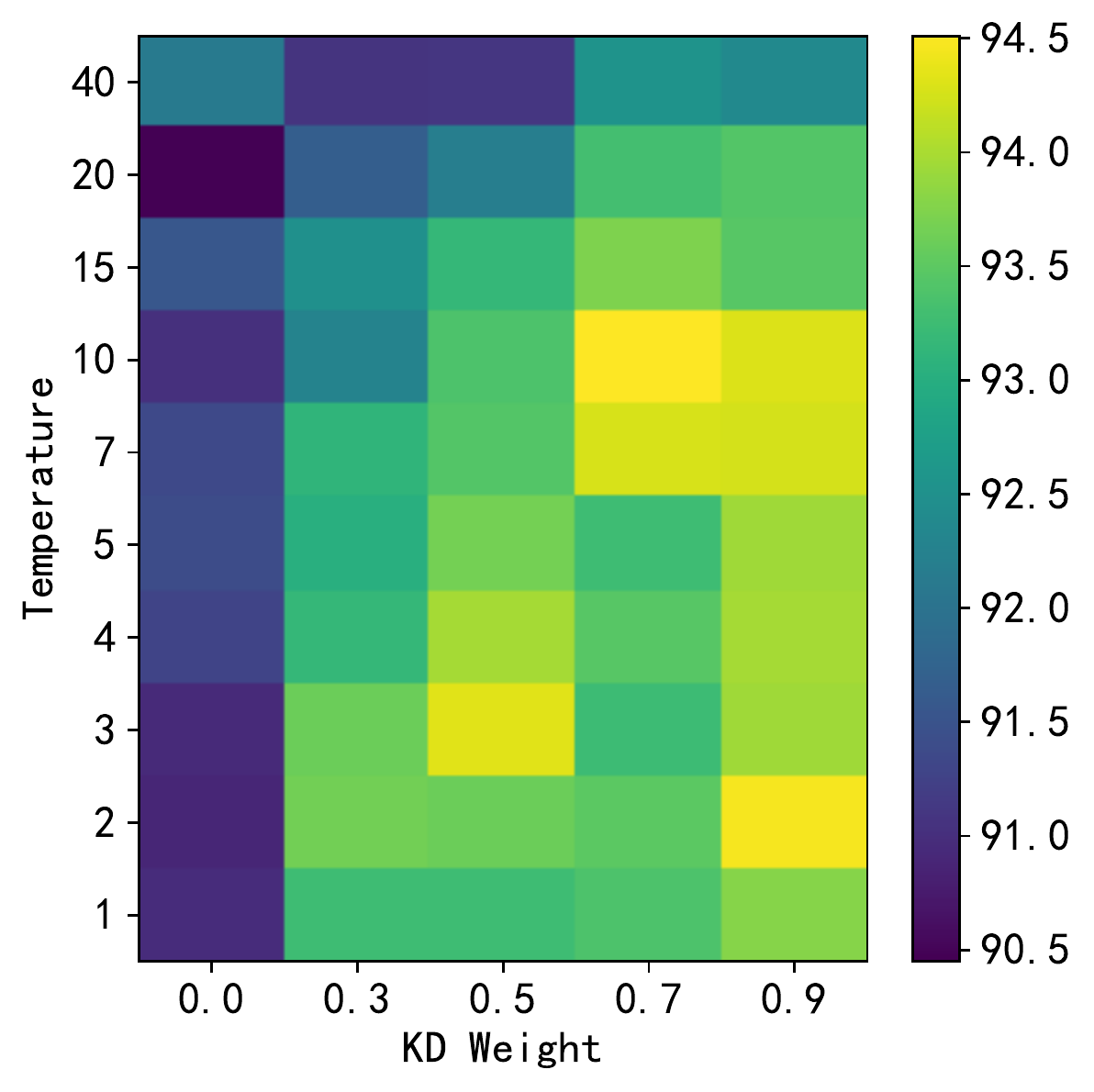}
\caption{Student model test accuracy for various temperature ($\tau$) and loss weight ($\alpha$) combinations. The model was sampled from a generator with random parameters, and trained with KD on CIFAR10 using the DARTS teacher.
The table suggests that there is a positive correlation between the KD loss weight and the performance of the student model. Note that the variability shown when the loss is set to 0 is solely due to the inherent stochasticity of the training process.
}
\label{fig:ablation_study}
\end{figure}

\end{document}